\documentclass[runningheads]{llncs}

\usepackage[mobile]{eccv}

\usepackage{eccvabbrv}

\usepackage{xcolor,colortbl}
\usepackage{multirow}
\usepackage{amssymb}

\usepackage{graphicx}
\usepackage{booktabs}
\usepackage{arydshln}

\usepackage[accsupp]{axessibility}  

\definecolor{best}{HTML}{fffeac}
\newcommand{\best}{\cellcolor{best}}
\definecolor{secondbest}{HTML}{ffacac}
\newcommand{\secondbest}{\cellcolor{secondbest}}
\DeclareMathOperator*{\argmin}{arg\,min}

\usepackage[pagebackref,breaklinks,colorlinks,citecolor=eccvblue]{hyperref}

\usepackage{orcidlink}

\begin{document}

\title{Optimizing Illuminant Estimation in Dual-Exposure HDR Imaging} 

\titlerunning{Illuminant Estimation in HDR Imaging}

\author{Mahmoud Afifi \and
Zhenhua Hu \and
Liang Liang}

\authorrunning{M Afifi, Z Hu, and L Liang}

\institute{Google}

\maketitle

\begin{abstract}
High dynamic range (HDR) imaging involves capturing a series of frames of the same scene, each with different exposure settings, to broaden the dynamic range of light. This can be achieved through burst capturing or using staggered HDR sensors that capture long and short exposures simultaneously in the camera image signal processor (ISP). Within camera ISP pipeline, illuminant estimation is a crucial step aiming to estimate the color of the global illuminant in the scene. This estimation is used in camera ISP white-balance module to remove undesirable color cast in the final image. Despite the multiple frames captured in the HDR pipeline, conventional illuminant estimation methods often rely only on a single frame of the scene. In this paper, we explore leveraging information from frames captured with different exposure times. Specifically, we introduce a simple feature extracted from dual-exposure images to guide illuminant estimators, referred to as the dual-exposure feature (DEF). To validate the efficiency of DEF, we employed two illuminant estimators using the proposed DEF: 1) a multilayer perceptron network (MLP), referred to as exposure-based MLP (EMLP), and 2) a modified version of the convolutional color constancy (CCC) to integrate our DEF, that we call ECCC. Both EMLP and ECCC achieve promising results, in some cases surpassing prior methods that require hundreds of thousands or millions of parameters, with only a few hundred parameters for EMLP and a few thousand parameters for ECCC.

\end{abstract}

\section{Introduction and Related Work}
\label{sec:intro}

\begin{figure}[t]
\centering
\includegraphics[width=\linewidth]{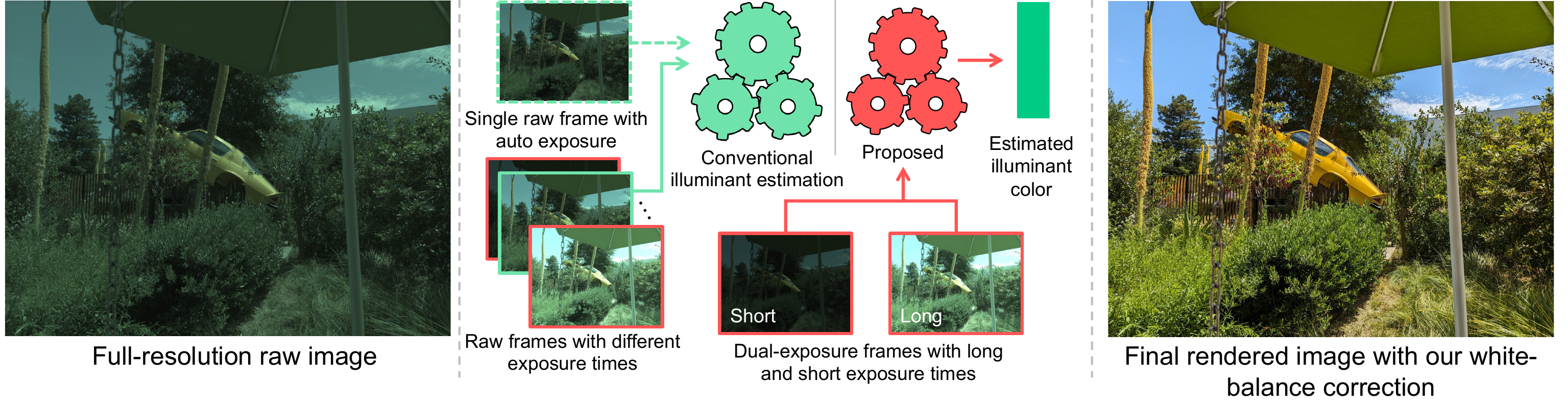}
\vspace{-3mm}
\caption{Conventional illuminant estimators often rely on a single frame for illuminant color estimation. Although the HDR camera pipeline includes at least two frames of the same scene, conventional methods usually consider only a single frame (either with auto exposure, shown in the dashed green line, merged frame, or one frame of the burst capture) \cite{HDR+, yahiaoui2019overview, delbracio2021mobile, MSCHAPTER}. This paper proposes leveraging information from dual-exposure capturing in multi-exposure HDR imaging to enhance illuminant estimation in camera pipelines. Our method uses frames captured at long and short exposures, available in multi-exposure bursts \cite{HDR-on-mobile} or staggered HDR sensors \cite{shdr1, shdr2}. Achieving comparable or superior results, our method employs lightweight models ($\sim$300--6000 parameters) compared to those using hundreds of thousands or millions of parameters. In this figure and the following figures, all raw images have the gamma operator applied to aid visualization, and all sRGB images are rendered using the HDR+ camera pipeline \cite{HDR+}.}
\vspace{-1mm}
\label{fig:teaser}
\end{figure}

Camera image signal processor (ISP) comprises several modules, each dedicated to enhancing specific aspects of the quality of captured raw images by the camera sensor \cite{delbracio2021mobile}. One key component among these modules is the white-balance module, which aims to eliminate undesirable color casts introduced by the combination of scene lighting and camera sensitivity. To achieve this, the auto white-balance module runs an illuminant estimator onboard the camera ISP to estimate the RGB color of the illuminant, under the assumption that a single global illuminant illuminates the scene for simplicity \cite{HDR+, gijsenij2011computational}. Image white balancing can thus be described as follows:

\begin{equation}
\label{eq:wb1}
I_{\texttt{WB}} = \text{diag}\left([\frac{\hat{\ell}_G}{\hat{\ell}_R}, 1, \frac{\hat{\ell}_G}{\hat{\ell}_B}]^T\right) I,
\end{equation}

\begin{equation}
\label{eq:wb2}
[\hat{\ell}_R, \hat{\ell}_G, \hat{\ell}_B]^T = f(I),
\end{equation}

\noindent where $I$ and $I_{\texttt{WB}}$ are $3 \times k$ RGB colors of raw image and white-balanced raw image, respectively, with $k$ refers to the total number of pixels in the image, $\text{diag}(.)$ creates a 3$\times$3 diagonal matrix from a given vector, $\hat{\ell} \in \mathbb{R}^3$ refers to the estimated illuminant color vector, $T$ refers to vector transpose, and $f$ is an illuminant estimator function.
  
Modern camera ISPs capture several frames of each scene with different exposure times to enhance final image dynamic range \cite{HDR-on-mobile, HDR-on-mobile2}. This can be achieved through burst capturing, involving rapid capture with varying exposure times in quick succession, or by using staggered high dynamic range (sHDR) sensors that simultaneously capture long and short exposures \cite{shdr1, shdr2}. The captured images are then combined, incorporating multiple exposures of the same scene at different levels, resulting in a greater dynamic range than what would be possible with a single image \cite{fusion0, fusion1, fusion2, le2023single}. This HDR imaging camera pipeline, therefore, includes more than a single frame of the captured scene, and internal camera ISP modules, such as the auto white-balance module, can access these additional frames. While such additional frames may have beneficial information to help illuminant estimators, conventional illuminant estimation methods often rely on a single frame (e.g., \cite{BMVC1, FFCC, FC4, BoCF, C4, TLCC, PCC}) in both traditional single-frame camera pipelines \cite{karaimer2016software} and multi-frame HDR ISPs \cite{HDR+}.

Such single-frame illuminant estimators can be categorized into: 1) statistical methods that rely on statistics computed from input raw image information (e.g., colors, edges) \cite{GW, SoG, NUS, GE, wGE, GI, Oguzhan} and 2) learning-based methods that learn, from a set of training images labeled with ground-truth illuminant colors, to map from input raw color information to the corresponding illuminant color \cite{brainard1997bayesian, Gamut1, bianco2012color, BMVC1, CCC, FC4, Seoung, APAP, CWCC, C4, CLCC, TLCC, PCC}. While the latter category typically achieves better results than the statistical methods, most learning-based methods are camera-dependent, meaning that they require domain adaptation or fine-tuning when deployed on new cameras to achieve similar accuracy on cameras used for training due to the influence of the camera response function on both raw image colors and ground-truth colors \cite{SIIE, hernandez2020multi, xiao2020multi, C5, afifi2021semi, Solomatov2023spectral}. 

A limited number of attempts have proposed learning-based methods that go beyond the single-frame input scheme. For example, the cross-camera convolutional color constancy (C5) \cite{C5} suggests utilizing additional unlabeled images captured by the testing camera, in addition to the primary single-frame image of the scene being captured to improve the generalization for cameras that were not included in the training phase. Xing et al. \cite{xing2022point} proposed the use of a depth map, captured by a time-of-flight (ToF) sensor, along with the primary raw image to predict illuminant color in the scene by leveraging the geometry information obtained from depth map. Abdelhamed et al., \cite{TWOCAMS} proposed leveraging the presence of two cameras in most modern mobile phone devices. Assuming dual streaming from both cameras, they derived a feature within the framework of chromagenic color constancy theory \cite{Chromagenic, Chromagenic1, Chromagenic2} to enhance the accuracy of illuminant estimation, yielding promising results. 

Our method, in contrast to the majority of prior work, adopts the strategy of benefiting from multiple frames available in the camera ISP (similar to \cite{TWOCAMS}). However, unlike \cite{TWOCAMS, xing2022point}, which requires streaming from dual cameras and may lead to impractical high power consumption, our method relies on two frames captured of the same scene under different exposure settings, already present in the HDR imaging pipeline, to estimate the illuminant color of the captured scene (see Fig. \ref{fig:teaser}). Specifically, we utilize: 1) a frame captured with short exposure time, that we refer to as short-exposure image ($I_s$), and 2) another frame of the same scene captured with long exposure time, that we refer to as long-exposure image ($I_l$). Having dual-exposure images in the HDR imaging pipeline is feasible, making our method practical for most HDR imaging pipelines. Accordingly, Eq. \ref{eq:wb2} can be modified to include our dual-exposure input images as follows:

\begin{equation}
\label{eq:wb3}
[\hat{\ell}_R, \hat{\ell}_G, \hat{\ell}_B]^T = f(I_l, I_s).
\end{equation}

Our objective in this work is to design the function $f$ in a manner that enables the effective utilization of the additional information provided by the dual-exposure images, $I_l$ and $I_s$.

\subsubsection{Contribution}
In this paper, we present a feature inspired by the chromagenic color constancy theory \cite{Chromagenic, Chromagenic1, Chromagenic2}, termed the dual-exposure feature (DEF), that is derived from images captured with both short exposure time ($I_s$) and long exposure time ($I_l$). DEF leverages the variations in chromatic information between these dual-exposure images, providing valuable guidance for illuminant estimator methods. To assess its effectiveness, we trained a lightweight multilayer perceptron network (MLP) for illuminant estimation that utilizes solely our DEF as input, departing from the conventional approach of using actual RGB values of the captured scene colors. Additionally, we explored the integration of DEF into an established color constancy framework, specifically the convolutional color constancy (CCC) \cite{hubel2007white, CCC, FFCC, C5}, which we refer to as exposure-based CCC (ECCC). The experimental results on a multi-exposure dataset, collected to evaluate our work empirically, show that these models -- namely, EMLP and ECCC -- achieve promising results with a reasonable number of parameters -- 354 learnable parameters for EMLP and 6,156 learnable parameters for ECCC. This outperformance across diverse evaluation metrics  is observed when comparing to prior methods that require significantly higher number of parameters, ranging from tens to hundreds of thousands, or even millions.

\section{Illuminant-Linked Dual Exposure Feature}
\label{sec:def}

To develop an efficient illuminant estimator that benefits from both $I_l$ and $I_s$, we introduce a compact feature that aims to capture the correlation between these dual-exposure images. Our feature is inspired by the chromagenic color constancy theory \cite{Chromagenic, Chromagenic1, Chromagenic2}. Specifically, we explore the analogy between long and short exposure images, $I_l$ and $I_s$, and aligned images captured by two cameras \cite{TWOCAMS} under the chromagenic color constancy theory, leading us to develop our dual-exposure feature (DEF). To begin, we review the chromagenic color constancy theory under the Lambertian reflectance model with a single illuminant assumption. The captured raw image $I$ can mathematically be described by:
\begin{equation}
\label{eq:image}
I_\rho^{(y)} = \int_\gamma S_{\rho}(x)  D(x)  R(y, x)\, dx + z,
\end{equation}
\noindent where $S\left(\cdot\right)$, $D\left(\cdot\right)$, and $R\left(\cdot\right)$ represent the camera response function (typically represented by camera sensitivity, infrared cut-off filter, and spectral lens transmission), the spectral power distribution of light, and scene reflectance, respectively. Here, $x$ refers to a wavelength within the visible range $\gamma$, $y$ refers to the pixel location in image $I$, $\rho \in \{R, G, B\}$ refers to the color channel, and $z$ denotes the undesired noise, typically represented by signal-dependent and additive components \cite{abdelhamed2018high}. According to the chromagenic color constancy theory \cite{Chromagenic2}, if we capture two images of the same scene with a specially chosen chromagenic filter, $Q$, applied between image captures, the linear color matrix that maps between those captured images is \textit{unique} to the illuminant color present in this scene. That is, given an image, $I$, that is captured by the main camera, and a filtered image, $I_f$, that can be described as:
\begin{equation}
\label{eq:filtered_image}
I_{f_\rho}^{(y)} = \int_\gamma S_\rho(x) Q_\rho(x) D(x) R(y, x)\, dx + z,
\end{equation}
\noindent the 3$\times$3 color matrix $C_c$ that maps between the colors of $I$ and $I_f$ is indicative of the scene illuminant. This color matrix can be computed by minimizing the following equation: 

\begin{equation}
\label{eq:color_matrix}
\argmin_{C_c} \left\|C_c I_f - I\right\|_F,
\end{equation}

\noindent where $\left\|\cdot\right\|_F$ is the Frobenius norm. The closed-form solution for Eq. \ref{eq:color_matrix} can be obtained using the pseudoinverse (i.e., $C_c = I I_f^{\dagger}$). While theoretically validated, the chromagenic filter conditions required to obtain a \textit{unique} mapping matrix per illuminant are challenging to meet in practice. Empirically, Finlayson et al., \cite{Chromagenic} found that most filters exhibit a reasonable level of \textit{correlation} between the computed matrix and the illuminant color, excluding neutral filters, which consistently result in a scaling relationship between the unfiltered image, $I$, and the filtered image, $I_f$. Relaxing the conditions to include \textit{normal} colored filter, Abdelhamed et al., \cite{TWOCAMS} proposed using two cameras, with the second camera serving as the main camera after applying a colored filter---i.e., $S(\cdot) Q(\cdot)$. Surprisingly, even when chromagenic filter conditions are not met, using a normal colored filter \cite{Chromagenic} or another camera with a different response function \cite{TWOCAMS} still shows a correlation between the computed matrix $C_c$ and the illuminant color to some extent, achieving promising results.

We argue that this correlation arises due to the variations, or what can be considered a form of ``distortion'', in the colors captured by the second (filtered) camera when compared to the original colors captured by the main camera. This color distortion varies based on the interplay between the scene irradiance, $D(\cdot) R(\cdot)$,  and the camera response functions -- namely, the main camera's $S(\cdot)$ and the filtered/second camera's $S(\cdot) Q(\cdot)$ in Eqs. \ref{eq:image} and \ref{eq:filtered_image}. In the context of machine learning (ML) models, the camera response functions remain fixed across the entire dataset. Consequently, ML models, such as the one proposed in \cite{TWOCAMS}, learn the correlation between the ground-truth illuminant color and the scene irradiance through the matrix $C_c$ that represents the level of ``distortion'' between  the colors of captured scene images.

\begin{figure}[t]
\centering
\includegraphics[width=\linewidth]{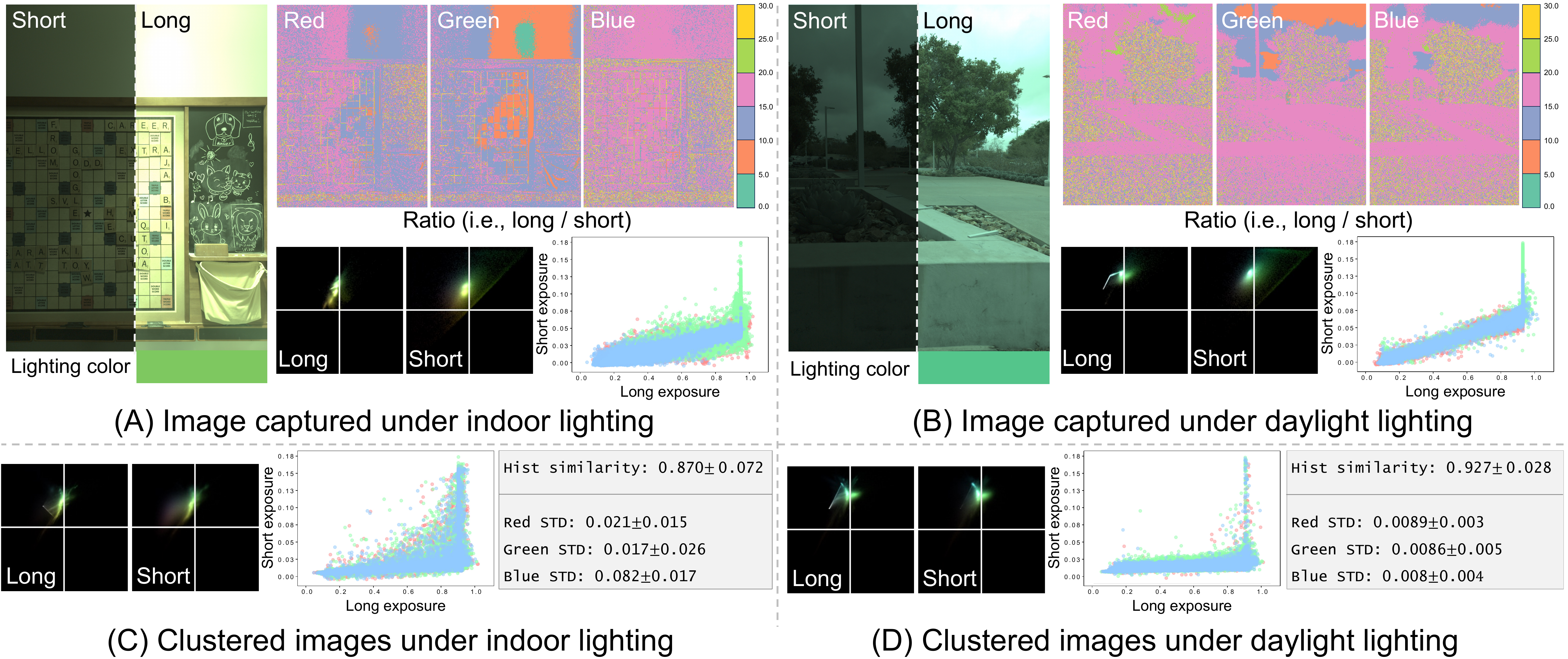}
\vspace{-3mm}
\caption{Cameras perceive different amount of photons when capturing the same scene with different exposure times. Images taken with both long and short exposure times exhibit variations in each channel due to the camera response function and scene irradiance. Additionally, spatial variations, in each color channel, can be observed based on object reflectance, as the interplay of lighting, object reflectance, and the camera response function leads to different outcomes. (A) and (B) show raw images of scenes captured under indoor and outdoor lighting, respectively. In (C) and (D), we present the average $rg$-chromaticity histogram and aggregated red, green, and blue pixel values from 25 images sharing similar lighting conditions in (A) and (B), respectively.}
\vspace{-1mm}
\label{fig:def}
\end{figure}

In a dual-exposure setup, we have a long-exposure image, $I_l$, and a short-exposure image, $I_s$, both capturing the same scene. The extent of color distortion in each image varies based on the scene irradiance; for instance, $I_s$ may exhibit a higher level of color distortion and noise than $I_l$ under suboptimal lighting conditions (e.g., indoor light), while $I_l$ may have a higher level of color distortion and over-saturated colors than $I_s$ in well-lit scenes (e.g., outdoor light). This difference in color distortion arises because the camera response function receives a different number of photons to form each image colors based on the exposure time and scene irradiance \cite{granados2010optimal}. As a result, $I_l$ and $I_s$ can exhibit color variations that differ across individual color channels \cite{HDR-radiance}, that are somewhat akin to those caused by a color filter (though to a lesser extent). Even within the same color channel, different levels of color differences between $I_l$ and $I_s$ can be observed spatially due to the interaction between the camera response function of that channel and the spatially varying scene irradiance (see Fig. \ref{fig:def}).

In Fig. \ref{fig:def}, we show two examples of dual-exposure images captured under different lighting conditions. As can be seen, the differences between the colors in $I_l$ and $I_s$ under each lighting condition exhibit variations, which are observable in the histogram similarity (here we use the Bhattacharyya distance similarity metric) and variations of the ratios in each color channel between $I_l$ and $I_s$. That is, the correlation between $I_l$ and $I_s$ \textit{is not} always a proportional scaling, as in the case of a neutral filter. We observed similar patterns as shown in Fig. \ref{fig:def} when studying examples from the two-camera dataset in \cite{TWOCAMS} (see Appendix \ref{sec:supp_discussion}).

Based on this discussion, we propose to compute the color matrix $C_c$ to map between the $rgb$-chromaticity values (i.e., $[R/\kappa, G/\kappa, B/\kappa]^T$, with $\kappa = R + G + B$) of $I_s$ and $I_l$. The reason for not using the RGB triples, similar to chromagenic color constancy, is that we aim to reduce the influence of intensity differences between $I_l$ and $I_s$ when computing $C_c$. In addition, we compute the covariance matrix, $C_v$, of the ratio image $X \in \mathbb{R}^{3\times k}$, where $X^{(y)}_{\rho} = I^{(y)}_{s_\rho} / \left(I^{(y)}_{l_\rho} + \epsilon\right)$, and $\epsilon$ is a small number added for numerical stability. Computing $C_v$ is performed as described in Eq. \ref{eq:cov} to measure the variance in each color channel between $I_s$ and $I_l$ and the joint variability across channels.

\begin{equation}
\label{eq:cov}
C_v(X) = \mathbb{E}\left[(X - \mathbb{E}[X])(X - \mathbb{E}[X])^T\right],
\end{equation}

\noindent where $\mathbb{E}\left(\cdot\right)$ is the expected value (mean) of the matrix. Both $C_c$ and $C_v$ form our dual-exposure feature (DEF) that represents the differences in color distortion in $I_l$ and $I_s$. Our DEF is represented as a vector $\in \mathbb{R}^{15}$, where we exclude the redundant values in $C_v$ over the symmetric positions. To evaluate DEF's effectiveness as a clue for scene illuminant colors, we employed a lightweight MLP for scene illuminant estimation (Fig. \ref{fig:main}-A), that we call exposure-based MLP (EMLP). EMLP relies solely on our DEF as input. It comprises an input fully connected (fc) layer with 9 output neurons, followed by leaky ReLU (LReLU) \cite{xu2015empirical}, two hidden fc layers, each with 9 output neurons with LReLU in between, and an output fc layer with three neurons. EMLP achieves results comparable to complex models with thousands or millions of parameters (refer to Tables \ref{tab:results-val} and \ref{tab:results-test}), while maintaining a lightweight design with just a few hundred parameters. Consequently, DEF is empirically shown as a valuable feature providing strong insights into the illuminant scene color.

\section{Integration with Convolutional Color Constancy}
\label{sec:eccc}
\begin{figure}[tb]
\centering
\includegraphics[width=\linewidth]{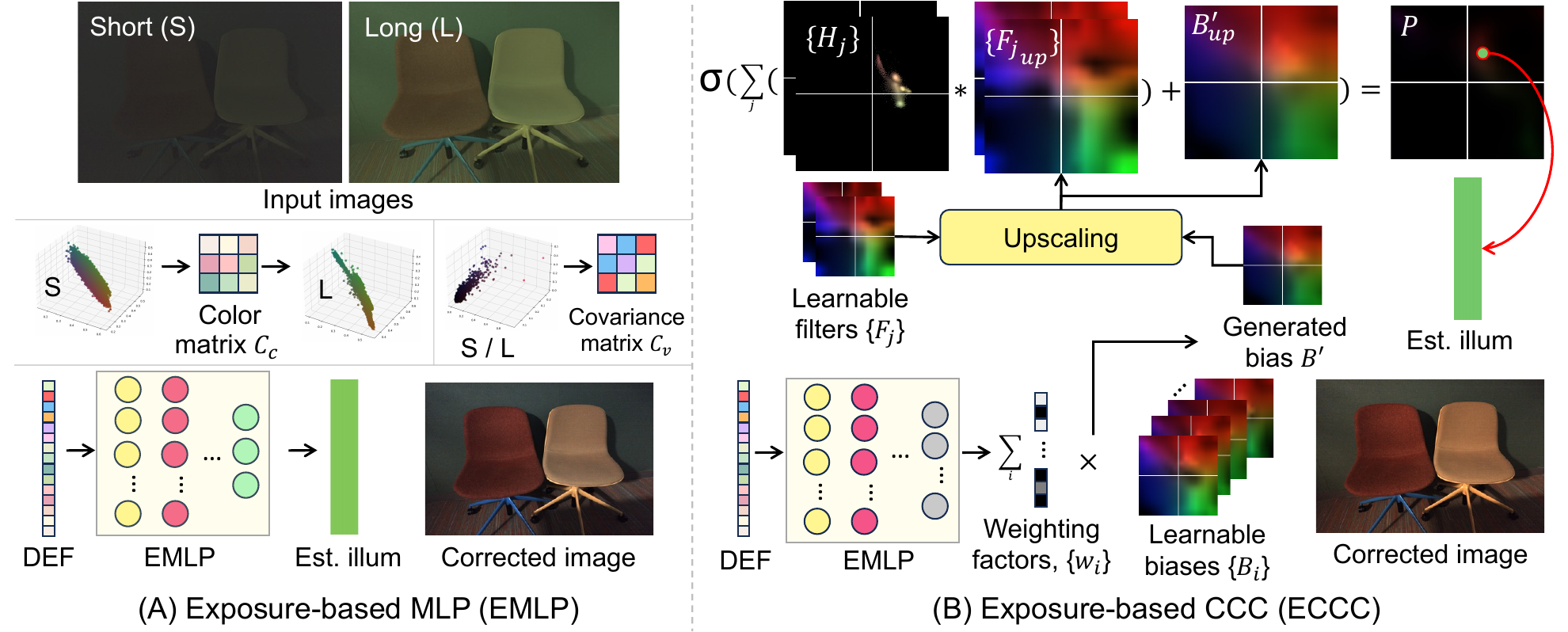}
\vspace{-3mm}
\caption{We present an illuminant-related dual-exposure feature (DEF), derived from a pair of images captured with short and long exposures. Using DEF, we deploy a simple multilayer perceptron network (MLP) with only 354 parameters, referred to as exposure-based MLP or EMLP, for illuminant estimation, as shown in (A). We further explore the integration of DEF into the CCC framework, as shown in (B), by dynamically generating bias map based on DEF. We denote this modified CCC framework as exposure-based CCC or ECCC.}
\vspace{-1mm}
\label{fig:main}
\end{figure}

In this section, we incorporate the proposed DEF into one of the most established frameworks for illuminant estimation. Specifically, we introduce modifications to the convolutional color constancy (CCC) framework \cite{hubel2007white, CCC, FFCC, C5} to leverage the benefits of the DEF, which we refer to as exposure-based CCC or ECCC for short. It is important to note that ECCC \textit{does not} introduce a new CCC method; instead, it serves as an illustration of how the DEF can be seamlessly integrated into existing, well-established illuminant estimation frameworks. 

Let's begin with a brief description of the CCC \cite{CCC, FFCC}. Given a single raw image, the CCC operates by learning one or more convolutional 2D filters, denoted as $\{F_j\}$, which convolve over the 2D histogram(s), $\{H_j\}$, of the image colors in the $uv$ color space (i.e., the log of $G/R$ and $G/B$ chroma values of pixel colors) \cite{finlayson2001color, CCC}. This convolutional operation can be accelerated by FFTs as proposed in \cite{FFCC}. Afterwards, a 2D bias map, $B$, is added, followed by a softmax operation, $\sigma$, to compute the ``probability'' map, $P$, of the illuminant bin in the $uv$ 2D histogram space. This simplified version of the CCC can be described as follows:

\begin{equation}
\label{eq:ccc}
P = \sigma\left(\sum_j{\left(F_j * H_j\right)} + B\right).
\end{equation}

In CCC \cite{CCC, FFCC}, the filters and bias are learned across the entire training dataset. This means it consists of a single bias and one or more filters (the number depends on the histograms used, either a single histogram for image colors or two histograms, including the edge color histogram). Later, C5 \cite{C5} proposes a hypernetwork that dynamically generates filters and bias based on the input raw image and additional images taken from the same camera, aiming to improve the generalization of CCC across cameras. While cross-camera color constancy is out of the scope of this paper, we propose to use our DEF to dynamically ``generate'' a bias map based on the input image. That is, we learn a bank of biases, $\{B_i\}$, where $i\in\{1, ..., n\}$, such that we linearly interpolate between them based on blending weights emitted from an MLP network that processes our DEF. In this way, the DEF controls the bias of the CCC model, acting as ``candidate'' illuminant priors within the $uv$ space for the input image (see Fig. \ref{fig:main}-B). To implement this, we need to have multiple bias maps, which definitely will lead to an impractical increase in model size. Thus, we propose to use a downsampled size (1/4 of histogram size) for the learnable biases, $\{B_i\}$. Furthermore, we propose to feed two histograms (i.e., $j \in \{l, s\}$) of both $I_l$ and $I_s$, denoted as $H_l$ and $H_s$, respectively, into the model. This leads to the learning of two downsampled filters, \(\{F_j\}\), corresponding to the histograms of the long-exposure image ($I_l$) and the short-exposure image ($I_s$). Learning small-sized filters and biases has the following benefits. First, we can learn many biases within an affordable model size (e.g., $n=20$ requires $\sim$6K parameters, while FFCC \cite{FFCC} with a single bias requires $\sim$12K parameters). Second, it implicitly produces smooth filters and biases, which are desirable to avoid overfitting \cite{FFCC, C5}. With this modification, Eq. \ref{eq:ccc} can be rewritten as follows: 

\begin{equation}
\label{eq:eccc1}
P = \sigma\left(\sum_j{\left(\uparrow(F_j) * H_j\right)} + B'_{\text{up}}\right),
\end{equation}

\begin{equation}
\label{eq:eccc2}
B'_{\text{up}} = \uparrow(\sum_{i=1}^{n}{w_i B_i}).
\end{equation}

\noindent where $\uparrow\left(\cdot\right)$ refers to upscaling through bilinear interpolation, $[w_1, ..., w_n]^T$ is a weighting vector produced by a lightweight MLP (similar to EMLP, but with $n$ output neurons) that processes our DEF.

After computing $P$, the estiamted illuminant can be obtained by: 

\begin{equation}
\label{eq:ccc3}
\hat{\ell}_u = \sum_{u, v} u P(u, v), \text{   } \hat{\ell}_v = \sum_{u, v} v P(u, v), 
\end{equation}

\begin{equation}
\label{eq:ccc4}
\hat{\ell} = \left[\exp{\left(-\hat{\ell_u}\right)}/q, 1/q, \exp{\left(-\hat{\ell_v}\right)}/q\right]^T,
\end{equation}

\begin{equation}
\label{eq:ccc5}
q = \sqrt{\exp{\left(-\hat{\ell_u}\right)^2} + \exp{\left(-\hat{\ell_v}\right)^2} + 1}.
\end{equation}

\section{Experiments}
\label{sec:experiments}

\subsection{Data}
\label{sec:data}

\begin{figure}[t]
\centering
\includegraphics[width=\linewidth]{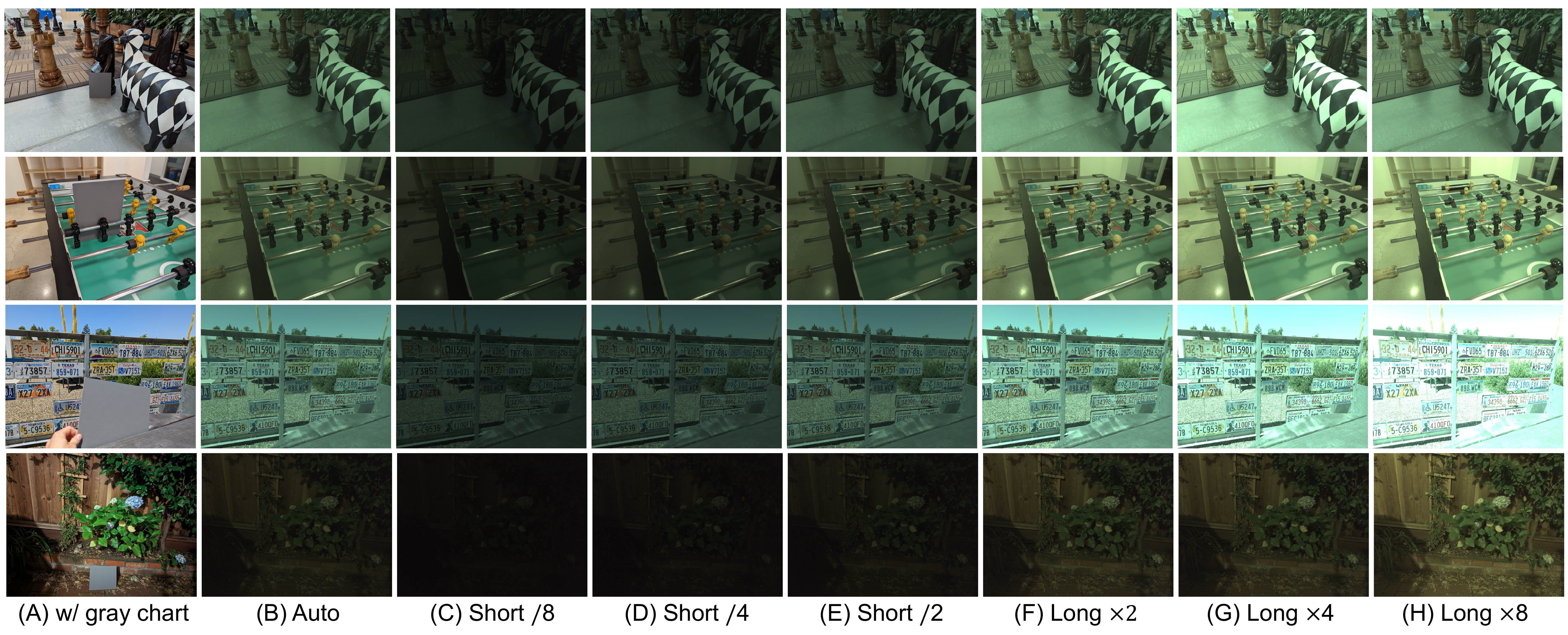}
\vspace{-3mm}
\caption{Examples from the dataset used in this work. For each scene, we captured the scene with a gray calibration object placed in the scene to obtain the ground-truth illuminant (A) and captured the scene using different exposure settings without the gray object (B-H). The terms `short $/ e$' (C-E) and `long $\times e$' (F-H) refer to multiplying and dividing auto exposure time by a factor $e$, respectively. The first image in (A) is displayed in sRGB, while the rest are shown in raw RGB space.}
\label{fig:dataset}
\vspace{-1mm}
\end{figure}

To the best of our knowledge, the existing HDR multi-exposure datasets (e.g., \cite{endoSA2017, hanji2022sihdr}) do not provide ground-truth illuminant colors for training and evaluation of our method. Thus, we compiled a dataset of scenes captured with both auto exposure and various multiple-exposure settings for each scene using Pixel 7 Pro camera (see Fig. \ref{fig:dataset}; for additional examples from the dataset, see Appendix \ref{sec:data-supp}). Images captured with auto exposure are used for training (in the case of learning-based methods) and evaluating other methods, while those captured with multiple exposure settings are employed for validating our method. Our set comprises 558 scenes, each captured using auto exposure and six additional exposure settings: $\times$2, $\times$4, $\times$8, 1/2, 1/4, and 1/8, indicating adjustments to the original exposure time. Specifically, $\times$2 signifies doubling the original exposure time, $\times$4 and $\times$8 denote quadrupling and octupling, respectively, while 1/2, 1/4, and 1/8 represent dividing the original exposure time by 2, 4, and 8, respectively. These adjustments are dynamically computed as ratios from the initial auto exposure, following the common practice in multi-exposure HDR imaging on smartphone cameras \cite{HDR-on-mobile, HDR-on-mobile2}. 

The camera was fixed on a tripod to minimize potential misalignment between images captured by different exposure settings. In total, we captured 3,906 raw images (558$\times$7 captures of each scene). The dataset is organized as follows: 83 and 86 scenes were randomly selected for validation and testing, respectively, with the remaining 387 scenes were used for training. Additionally, each scene was captured with a gray calibration object to obtain the ground-truth illuminant. The images with the gray calibration object are not included in any of the training, validation, nor testing sets. 

During training, we optionally augment training data using chromatic adaptation. Specifically, we employ clustering on the training data using our DEF as query feature. We use K-means \cite{KMEAN} with L2 to create 80 clusters. Subsequently, we augment the training data by generating three additional images for each sample. For each augmented image, we randomly select an illuminant color from the cluster to which the original image belongs. We then apply Von Kries transform \cite{fairchild2013color} to remove the original illuminant from the image, using the actual ground truth, and apply the newly selected illuminant color to the image using the diagonal scaling operator. This new illuminant serves as the ground truth for the augmented image.

As our method is designed to handle two images, namely $I_s$ and $I_l$, we use, without loss of generality, a single exposure factor, $e$, for both images, such that $I_s$ was captured with $1/e$ of the auto-exposure time, while $I_l$ was captured with $\times e$ of the auto-exposure time, for simplicity. In Sec. \ref{sec:resutls}, we present an ablation study on the impact of different values of $e$ on our final results.

\subsection{Training}
\label{sec:training}
We trained EMLP and ECCC using the Adam optimizer \cite{kingma2014adam} with a mini-batch size set to 32 for 1000 and 200 epochs, respectively. For EMLP, the learning rate was $10^{-3}$, while ECCC was trained with an incremental mini-batch size (similar to \cite{C5}) with a learning rate of $5\times10^{-3}$ with a cosine annealing schedule \cite{loshchilov2016sgdr}, and the weight decay (i.e., L2 regularization penalty) was set to $10^{-5}$. For ECCC, the filter weights were initialized to zeros, and the biases were initialized to $n$ 2D histograms of training ground-truth illuminant colors after clustering into $n$ clusters using our DEF as a query feature with K-means. The $n$ histograms of training ground-truth illuminants were first processed by morphological dilation using a 3$\times$3 diamond-shaped structuring element before being used as initial values for our learnable biases. This initialization improves the results, as it assists in establishing reasonable ``candidate'' illuminant priors linked to the corresponding DEFs (see Sec. \ref{sec:resutls} for an ablation study).

Both models were trained using the angular error, $L\left(\cdot\right)$, between the predicted illuminant color $\hat{\ell}$ and the ground-truth illuminant $\ell$, as described in the following equation \cite{finlayson1995color}:

\begin{equation}
\label{eq:ae}
L(\ell, \hat{\ell}) = \text{cos}^{-1}\left(\frac{\ell \cdot \hat{\ell}}{\left\|\ell\right\|\left\|\hat{\ell}\right\|}\right),
\end{equation}

\noindent where $\left\|\cdot\right\|$ denotes the Euclidean norm, and ($\cdot$) represents the vector dot product. For ECCC, we further added two smoothness loss terms to encourage smoothness in the learned filters and biases, similar to \cite{C5}. See Appendix \ref{sec:eccc-supp} for more details.

\subsection{Results}
\label{sec:resutls}
We conducted a comprehensive comparison of our proposed methods against various existing techniques, which include: training-free statistical methods \cite{GW, SoG, NUS, GI}, camera-independent learning-based methods \cite{SIIE, C5}, and camera-dependent learning-based methods \cite{GAMUT, FFCC, APAP, BoCF, C4, CWCC, TLCC, PCC}. For the camera-dependent learning-based methods, we trained each model on our data using the provided code and recommended parameters by the respective authors. Meanwhile, for sensor-independent learning methods, we utilized the provided pre-trained models on other datasets (e.g., \cite{NUS, Cube+, gehler2008bayesian}).

For the sake of conducting realistic experiments, we assessed all methods, including ours, on 384$\times$256 images---a suitable size for evaluating illuminant estimation methods designed for embedded hardware devices; such devices may utilize even smaller image sizes in white-balance modules \cite{FFCC}. For CCC methods \cite{FFCC, C5}, including the  ECCC, 64$\times$64 histograms were used.

The ECCC incorporates the proposed DEF into the CCC by generating an interpolated bias map based on the DEF. This can be seen as a spatial case analogous to C5 \cite{C5}, where C5 utilizes a neural network to emit bias and convolution filters based on additional images taken by the same camera to enhance generalization across cameras. To validate our modification against C5, we trained different versions of C5 not aimed at improving cross-camera generalization (as our method also focuses on a single camera). We refer to these modified versions as follows: 
\begin{itemize}
\item \textbf{C5 (model A):} We use the histogram of the averaged short and long exposure images without additional images. That is, the C5 network uses a single histogram to generate CCC model parameters. 
\item \textbf{C5 (model B):} We use histograms of both images taken by long and short exposures as input to the C5 network to generate CCC model parameters. The CCC mode is then applied to the histogram of the averaged long-short images.
\item \textbf{C5 (models C and D):} Both Model C and Model D use histograms of both images taken by long and short exposures to feed the C5 network and generate CCC model that is then applied to histogram of short exposure image (model C) and long exposure image (model D).
\end{itemize}

\begin{table}[!b]
\centering
\caption{Angular errors on the validation set. $\circledcirc$ and $\lozenge$ denote camera-independent models and training-free statistical methods, respectively. Methods are listed chronologically by publication year, with the top and second-best results highlighted in yellow and red. We present the results of camera-dependent trained models for various versions of C5 \cite{C5}, identified as models A, B, C, and D (see main text for more details). Our results are reported using EMLP, ECCC, and the ensemble model (EMLP + ECCC). Ablation studies are included using different mapping matrices (CM: $3\times3$ color mapping matrix, TM: $3\times4$ affine transformation matrix, HM: $3\times3$ homography matrix), different color representations (rgb: raw RGB triplet as used in \cite{Chromagenic, Chromagenic1, Chromagenic2, Chromagenic3, TWOCAMS}, $rg$/$rgb$-chroma: $rg$-chromaticity/$rgb$-chromaticity of the raw RGB triplet), different exposure ratios ($e \in \{2, 4, 8\}$), different input image sizes and histograms ($s$ indicates 48$\times$32 input images, $s^2$ indicates 48$\times$32 input images and 32$\times$32 histograms), using different input histograms for ECCC of both long and short exposure images, average image, and long/short image, different number of learned biases ($n$), and results with (w/) and without (w/o) augmentation, w/ and w/o the covariance matrix's parameters (for EMLP), w/ and w/o DEF and bias initialization (BI) (for ECCC).}
\label{tab:results-val}
\scalebox{0.62}{
\begin{tabular}{|l|ccccccc|c|}
\hline
\multirow{2}{*}{\textbf{Method}} & \multicolumn{7}{c|}{\textbf{Validation set}} &  \multirow{2}{*}{\textbf{Params}} \\ \cline{2-8}
 & \multicolumn{1}{c|}{\textbf{Mean}} & \multicolumn{1}{c|}{\textbf{Med.}} & \multicolumn{1}{c|}{\textbf{Tri.}} & \multicolumn{1}{c|}{\begin{tabular}[c]{@{}c@{}}\textbf{Best} \\ \textbf{25}\%\end{tabular}} & \multicolumn{1}{c|}{\begin{tabular}[c]{@{}c@{}}\textbf{Worst} \\ \textbf{25}\%\end{tabular}} & \multicolumn{1}{c|}{\begin{tabular}[c]{@{}c@{}}\textbf{Worst} \\ \textbf{5}\%\end{tabular}} & \textbf{Max} &   \\ \hline

Grayworld$^\lozenge$ \cite{GW} & \multicolumn{1}{c|}{5.54} & \multicolumn{1}{c|}{3.13} & \multicolumn{1}{c|}{4.11} & \multicolumn{1}{c|}{0.97} & \multicolumn{1}{c|}{12.83} & \multicolumn{1}{c|}{19.45} & 28.82 &  -  \\ \hline

Shades of Gray$^\lozenge$ \cite{SoG} & \multicolumn{1}{c|}{7.16} & \multicolumn{1}{c|}{6.21} & \multicolumn{1}{c|}{6.43} & \multicolumn{1}{c|}{0.99} & \multicolumn{1}{c|}{15.06} & \multicolumn{1}{c|}{18.39} & 19.21 &  - \\ \hline

PCA$^\lozenge$ \cite{NUS} & \multicolumn{1}{c|}{6.13} & \multicolumn{1}{c|}{4.53} & \multicolumn{1}{c|}{5.02} & \multicolumn{1}{c|}{0.91} & \multicolumn{1}{c|}{13.66} & \multicolumn{1}{c|}{18.28} & 20.33 &  - \\ \hline

Gamut (pixels) \cite{GAMUT}  & \multicolumn{1}{c|}{7.53} & \multicolumn{1}{c|}{7.38} & \multicolumn{1}{c|}{7.03} & \multicolumn{1}{c|}{2.04} & \multicolumn{1}{c|}{13.96} & \multicolumn{1}{c|}{18.34} & 21.01 & 636  \\ \hline

Gamut (edges) \cite{GAMUT}  & \multicolumn{1}{c|}{7.03} & \multicolumn{1}{c|}{6.06} & \multicolumn{1}{c|}{6.16} & \multicolumn{1}{c|}{1.49} & \multicolumn{1}{c|}{14.54} & \multicolumn{1}{c|}{19.51} & 21.98 &  324 \\ \hline

FFCC \cite{FFCC}  & \multicolumn{1}{c|}{3.77} & \multicolumn{1}{c|}{\best 1.73} & \multicolumn{1}{c|}{\best 2.32} & \multicolumn{1}{c|}{\best0.54} & \multicolumn{1}{c|}{9.56} & \multicolumn{1}{c|}{16.08} & 28.99 & 12,288 \\ \hline

Gray Index$^\lozenge$ \cite{GI} & \multicolumn{1}{c|}{6.00} & \multicolumn{1}{c|}{3.62} & \multicolumn{1}{c|}{4.28} & \multicolumn{1}{c|}{\secondbest 0.58} & \multicolumn{1}{c|}{15.44} & \multicolumn{1}{c|}{24.91} & 31.72 &  -  \\ \hline

APAP-LUT \cite{APAP}  & \multicolumn{1}{c|}{4.84} & \multicolumn{1}{c|}{3.30} & \multicolumn{1}{c|}{3.87} & \multicolumn{1}{c|}{1.12} & \multicolumn{1}{c|}{10.84} & \multicolumn{1}{c|}{16.92} &  27.53 & \secondbest 289\\ \hline

SIIE$^\circledcirc$ \cite{SIIE}  & \multicolumn{1}{c|}{4.70} & \multicolumn{1}{c|}{4.04} & \multicolumn{1}{c|}{4.23} & \multicolumn{1}{c|}{1.35} & \multicolumn{1}{c|}{9.21} & \multicolumn{1}{c|}{13.27} & 16.24 & 1,008,044 \\ \hline

BoCF \cite{BoCF}  & \multicolumn{1}{c|}{4.84} & \multicolumn{1}{c|}{3.30} & \multicolumn{1}{c|}{3.87} & \multicolumn{1}{c|}{1.12} & \multicolumn{1}{c|}{10.84} & \multicolumn{1}{c|}{16.92} & 27.53 &  59,354 \\ \hline

C4 \cite{C4}  & \multicolumn{1}{c|}{3.91} & \multicolumn{1}{c|}{3.26} & \multicolumn{1}{c|}{3.24} & \multicolumn{1}{c|}{1.06} & \multicolumn{1}{c|}{8.50} & \multicolumn{1}{c|}{12.76} & 17.35 & 5,115,657\\ \hline

CWCC \cite{CWCC}  & \multicolumn{1}{c|}{4.49} & \multicolumn{1}{c|}{3.35} & \multicolumn{1}{c|}{3.48} & \multicolumn{1}{c|}{1.61} & \multicolumn{1}{c|}{9.21} & \multicolumn{1}{c|}{13.31} & 16.11 & 100,830 \\ \hline

C5$^\circledcirc$ \cite{C5}  & \multicolumn{1}{c|}{4.01} & \multicolumn{1}{c|}{2.81} & \multicolumn{1}{c|}{3.14} & \multicolumn{1}{c|}{1.08} & \multicolumn{1}{c|}{8.51} & \multicolumn{1}{c|}{12.47} & 21.81 & 411,711 \\ \hline

C5 (model A)  \cite{C5}  & \multicolumn{1}{c|}{3.93} & \multicolumn{1}{c|}{\secondbest 2.29} & \multicolumn{1}{c|}{2.90} & \multicolumn{1}{c|}{1.01} & \multicolumn{1}{c|}{9.40} & \multicolumn{1}{c|}{13.33} & 18.75  & 171,511 \\ \hline

C5 (model B)  \cite{C5}   & \multicolumn{1}{c|}{4.49} & \multicolumn{1}{c|}{2.84} & \multicolumn{1}{c|}{3.24} & \multicolumn{1}{c|}{0.88} & \multicolumn{1}{c|}{10.52} & \multicolumn{1}{c|}{15.23} & 20.55 & 213,831 \\ \hline

C5 (model C)  \cite{C5}  & \multicolumn{1}{c|}{4.25} & \multicolumn{1}{c|}{2.67} & \multicolumn{1}{c|}{3.19} & \multicolumn{1}{c|}{0.93} & \multicolumn{1}{c|}{10.12} & \multicolumn{1}{c|}{14.57} & 19.82 & 213,831 \\ \hline

C5 (model D)  \cite{C5}  & \multicolumn{1}{c|}{4.51} & \multicolumn{1}{c|}{2.94} & \multicolumn{1}{c|}{3.38} & \multicolumn{1}{c|}{0.87} & \multicolumn{1}{c|}{10.74} & \multicolumn{1}{c|}{16.01} & 20.74 & 213,831 \\ \hline

TLCC \cite{TLCC}  & \multicolumn{1}{c|}{4.16} & \multicolumn{1}{c|}{2.72} & \multicolumn{1}{c|}{3.15} & \multicolumn{1}{c|}{0.81} & \multicolumn{1}{c|}{9.35} & \multicolumn{1}{c|}{14.05} & 21.39 & 32,910,186 \\ \hline

PCC \cite{PCC}  & \multicolumn{1}{c|}{4.61} & \multicolumn{1}{c|}{4.19} & \multicolumn{1}{c|}{4.15} & \multicolumn{1}{c|}{1.75} & \multicolumn{1}{c|}{8.50} & \multicolumn{1}{c|}{12.06} & 17.86 & 450 \\ \hdashline

EMLP ($e=8$, $rg$-chroma, HM, w/o cov) & \multicolumn{1}{c|}{5.86} & \multicolumn{1}{c|}{4.26} & \multicolumn{1}{c|}{4.96} & \multicolumn{1}{c|}{1.36} & \multicolumn{1}{c|}{12.3} & \multicolumn{1}{c|}{16.36} & 20.44 & 300\\ \hline

EMLP ($e=8$, $rgb$-chroma, TM, w/o cov) & \multicolumn{1}{c|}{4.78} & \multicolumn{1}{c|}{3.94} & \multicolumn{1}{c|}{4.27} & \multicolumn{1}{c|}{1.14} & \multicolumn{1}{c|}{9.57} & \multicolumn{1}{c|}{13.20} & 16.54 & 363 \\ \hline

EMLP ($e=8$, rgb, CM, w/o cov) & \multicolumn{1}{c|}{4.54} & \multicolumn{1}{c|}{3.53} & \multicolumn{1}{c|}{3.87} & \multicolumn{1}{c|}{1.26} & \multicolumn{1}{c|}{9.51} & \multicolumn{1}{c|}{13.74} & 19.22 & 300\\ \hline

EMLP ($e=8$, $rg$-chroma, CM, w/o cov) & \multicolumn{1}{c|}{4.81} & \multicolumn{1}{c|}{3.98} & \multicolumn{1}{c|}{4.22} & \multicolumn{1}{c|}{1.29} & \multicolumn{1}{c|}{9.32} & \multicolumn{1}{c|}{12.75} & 14.99 & \best 255 \\ \hline

EMLP ($e=2$, $rgb$-chroma, CM, w/o cov) & \multicolumn{1}{c|}{4.35} & \multicolumn{1}{c|}{3.33} & \multicolumn{1}{c|}{3.57} & \multicolumn{1}{c|}{1.11} & \multicolumn{1}{c|}{9.44} & \multicolumn{1}{c|}{13.60} & 17.14 & 300 \\ \hline

EMLP ($e=4$, $rgb$-chroma, CM, w/o cov) & \multicolumn{1}{c|}{4.16} & \multicolumn{1}{c|}{3.56} & \multicolumn{1}{c|}{3.77} & \multicolumn{1}{c|}{1.00} & \multicolumn{1}{c|}{8.38} & \multicolumn{1}{c|}{12.16} & 16.79 & 300 \\ \hline

EMLP ($e=8$, $rgb$-chroma, CM, w/o cov) & \multicolumn{1}{c|}{4.07} & \multicolumn{1}{c|}{3.51} & \multicolumn{1}{c|}{3.64} & \multicolumn{1}{c|}{1.23} & \multicolumn{1}{c|}{\secondbest 7.77} & \multicolumn{1}{c|}{\best 10.38} & 13.29 &  300 \\ \hline

EMLP ($e=8$, $rgb$-chroma, CM, w/ cov) & \multicolumn{1}{c|}{3.77} & \multicolumn{1}{c|}{2.89} & \multicolumn{1}{c|}{2.92} & \multicolumn{1}{c|}{0.86} & \multicolumn{1}{c|}{8.59} & \multicolumn{1}{c|}{11.48} & 12.71 & 354 \\ \hline

EMLP-$s$ ($e=8$, $rgb$-chroma, CM, w/ aug, w/ cov) & \multicolumn{1}{c|}{3.83} & \multicolumn{1}{c|}{2.82} & \multicolumn{1}{c|}{3.27} & \multicolumn{1}{c|}{0.88} & \multicolumn{1}{c|}{8.23} & \multicolumn{1}{c|}{11.29} & 12.81 &  354\\ \hline

EMLP ($e=8$, $rgb$-chroma, CM, w/ aug, w/ cov) & \multicolumn{1}{c|}{\secondbest3.52} & \multicolumn{1}{c|}{2.43} & \multicolumn{1}{c|}{2.85} & \multicolumn{1}{c|}{0.86} & \multicolumn{1}{c|}{8.68} & \multicolumn{1}{c|}{11.32} & 13.11 & 354 \\ \hline


ECCC ($e=8$, $n=5$, w/ DEF, both) & \multicolumn{1}{c|}{3.69} & \multicolumn{1}{c|}{2.78} & \multicolumn{1}{c|}{3.11} & \multicolumn{1}{c|}{0.79} & \multicolumn{1}{c|}{7.83} & \multicolumn{1}{c|}{10.62} & 13.27 &  2,166 \\  \hline

ECCC ($e=8$, $n=10$, w/ DEF, both) & \multicolumn{1}{c|}{3.66} & \multicolumn{1}{c|}{2.69} & \multicolumn{1}{c|}{2.92} & \multicolumn{1}{c|}{0.95} & \multicolumn{1}{c|}{8.03} & \multicolumn{1}{c|}{11.06} & 12.32 & 3,496 \\ \hline

ECCC ($e=8$, $n=15$, w/ DEF, both) & \multicolumn{1}{c|}{3.58} & \multicolumn{1}{c|}{2.71} & \multicolumn{1}{c|}{2.91} & \multicolumn{1}{c|}{0.91} & \multicolumn{1}{c|}{7.96} & \multicolumn{1}{c|}{11.85} & 13.51 &  4,826 \\ \hline

ECCC ($e=2$, $n=20$, w/ DEF, both) & \multicolumn{1}{c|}{3.91} & \multicolumn{1}{c|}{2.99} & \multicolumn{1}{c|}{3.23} & \multicolumn{1}{c|}{1.07} & \multicolumn{1}{c|}{8.33} & \multicolumn{1}{c|}{11.84} & 15.23 &  6,156 \\ \hline

ECCC ($e=4$, $n=20$, w/ DEF, both) & \multicolumn{1}{c|}{4.28} & \multicolumn{1}{c|}{3.19} & \multicolumn{1}{c|}{3.36} & \multicolumn{1}{c|}{0.99} & \multicolumn{1}{c|}{9.63} & \multicolumn{1}{c|}{13.72} & 14.72 & 6,156\\ \hline

ECCC ($e=8$, $n=20$, w/o DEF, both) & \multicolumn{1}{c|}{4.02} & \multicolumn{1}{c|}{3.23} & \multicolumn{1}{c|}{3.69} & \multicolumn{1}{c|}{1.22} & \multicolumn{1}{c|}{9.17} & \multicolumn{1}{c|}{12.14} & 14.47 &  4,608 \\ \hline

ECCC ($e=8$, $n=20$, w/o BI, both) & \multicolumn{1}{c|}{3.99} & \multicolumn{1}{c|}{3.13} & \multicolumn{1}{c|}{3.51} & \multicolumn{1}{c|}{0.97} & \multicolumn{1}{c|}{8.53} & \multicolumn{1}{c|}{11.41} & 13.66 &  6,156 \\ \hline

ECCC ($e=8$, $n=20$, w/ DEF, avg) & \multicolumn{1}{c|}{4.03} & \multicolumn{1}{c|}{2.97} & \multicolumn{1}{c|}{3.34} & \multicolumn{1}{c|}{1.03} & \multicolumn{1}{c|}{8.68} & \multicolumn{1}{c|}{12.09} & 14.71 &  5,900 \\ \hline

ECCC ($e=8$, $n=20$, w/ DEF, short) & \multicolumn{1}{c|}{3.71} & \multicolumn{1}{c|}{2.74} & \multicolumn{1}{c|}{2.93} & \multicolumn{1}{c|}{1.00} & \multicolumn{1}{c|}{8.07} & \multicolumn{1}{c|}{12.20} & 15.43 &  5,900 \\ \hline

ECCC ($e=8$, $n=20$, w/ DEF, long) & \multicolumn{1}{c|}{3.66} & \multicolumn{1}{c|}{2.58} & \multicolumn{1}{c|}{2.85} & \multicolumn{1}{c|}{0.79} & \multicolumn{1}{c|}{8.20} & \multicolumn{1}{c|}{11.34} &  12.22 &  5,900 \\ \hline

ECCC ($e=8$, $n=20$, $rgb$-chroma, w/ DEF, both) & \multicolumn{1}{c|}{3.79} & \multicolumn{1}{c|}{2.73} & \multicolumn{1}{c|}{3.23} & \multicolumn{1}{c|}{0.82} & \multicolumn{1}{c|}{8.41} & \multicolumn{1}{c|}{11.63} & 15.44 & 6,156 \\ \hline

ECCC ($e=8$, $n=20$, w/ DEF, both, w/ aug) & \multicolumn{1}{c|}{3.61} & \multicolumn{1}{c|}{2.97} & \multicolumn{1}{c|}{3.14} & \multicolumn{1}{c|}{0.79} & \multicolumn{1}{c|}{\secondbest 7.77} & \multicolumn{1}{c|}{\secondbest 10.54} & \best 11.41 & 6,156 \\ \hline

ECCC-$s$ ($e=8$, $n=20$, w/ DEF, both) & \multicolumn{1}{c|}{3.69} & \multicolumn{1}{c|}{2.62} & \multicolumn{1}{c|}{3.07} & \multicolumn{1}{c|}{0.9} & \multicolumn{1}{c|}{8.12} & \multicolumn{1}{c|}{10.95} & 15.1 & 6,156 \\ \hline

ECCC-$s^2$ ($e=8$, $n=20$, w/ DEF, both) & \multicolumn{1}{c|}{4.04} & \multicolumn{1}{c|}{3.34} & \multicolumn{1}{c|}{3.50} & \multicolumn{1}{c|}{0.66} & \multicolumn{1}{c|}{8.75} & \multicolumn{1}{c|}{11.28} & 12.96 & 1,932 \\ \hline

ECCC ($e=8$, $n=20$, w/ DEF, both) & \multicolumn{1}{c|}{3.56} & \multicolumn{1}{c|}{2.50} & \multicolumn{1}{c|}{ 2.79} & \multicolumn{1}{c|}{0.87} & \multicolumn{1}{c|}{7.93} & \multicolumn{1}{c|}{11.32} & 13.47 & 6,156 \\ \hline

EMLP + ECCC & \multicolumn{1}{c|}{\best 3.24} & \multicolumn{1}{c|}{2.37 } & \multicolumn{1}{c|}{\secondbest 2.53} & \multicolumn{1}{c|}{ 0.84} & \multicolumn{1}{c|}{\best 7.11 } & \multicolumn{1}{c|}{ 10.64} & \secondbest 12.06 & 6,510 \\ \hline

\end{tabular}
}
\vspace{-2mm}
\end{table}

\begin{table}[!t]
\centering
\caption{Angular errors on the testing set. See caption of Table \ref{tab:results-val} for abbreviations.}
\label{tab:results-test}
\scalebox{0.63}{
\begin{tabular}{|l|ccccccc|}
\hline
\multirow{2}{*}{Method} & \multicolumn{7}{c|}{Testing set}  \\ \cline{2-8} 
 & \multicolumn{1}{c|}{Mean} & \multicolumn{1}{c|}{Med.} & \multicolumn{1}{c|}{Tri.} & \multicolumn{1}{c|}{\begin{tabular}[c]{@{}c@{}}Best \\ 25\%\end{tabular}} & \multicolumn{1}{c|}{\begin{tabular}[c]{@{}c@{}}Worst \\ 25 \%\end{tabular}} & \multicolumn{1}{c|}{\begin{tabular}[c]{@{}c@{}}Worst \\ 5\%\end{tabular}} & Max  \\ \hline

Grayworld$^\lozenge$ \cite{GW} & \multicolumn{1}{c|}{5.32} & \multicolumn{1}{c|}{3.70} & \multicolumn{1}{c|}{4.51} & \multicolumn{1}{c|}{0.95} & \multicolumn{1}{c|}{11.34} & \multicolumn{1}{c|}{14.41} & 17.75 \\ \hline

Shades of Gray$^\lozenge$ \cite{SoG} & \multicolumn{1}{c|}{6.77} & \multicolumn{1}{c|}{5.77} & \multicolumn{1}{c|}{6.08} & \multicolumn{1}{c|}{1.07} & \multicolumn{1}{c|}{14.24} & \multicolumn{1}{c|}{16.64} &  17.50 \\ \hline

PCA$^\lozenge$ \cite{NUS} & \multicolumn{1}{c|}{5.91} & \multicolumn{1}{c|}{4.99} & \multicolumn{1}{c|}{5.22} & \multicolumn{1}{c|}{1.07} & \multicolumn{1}{c|}{12.62} & \multicolumn{1}{c|}{17.10} & 19.08  \\ \hline

Gamut (pixels) \cite{GAMUT}  & \multicolumn{1}{c|}{7.49} & \multicolumn{1}{c|}{7.25} & \multicolumn{1}{c|}{7.37} & \multicolumn{1}{c|}{2.30} & \multicolumn{1}{c|}{12.96} & \multicolumn{1}{c|}{16.62} & 19.00 \\ \hline

Gamut (edges) \cite{GAMUT}  & \multicolumn{1}{c|}{5.90} & \multicolumn{1}{c|}{4.80} & \multicolumn{1}{c|}{4.93} & \multicolumn{1}{c|}{1.13} & \multicolumn{1}{c|}{12.20} & \multicolumn{1}{c|}{17.61} & 20.09  \\ \hline

FFCC \cite{FFCC}  & \multicolumn{1}{c|}{3.18} & \multicolumn{1}{c|}{\best 1.83} & \multicolumn{1}{c|}{\best 2.28} & \multicolumn{1}{c|}{\best 0.39} & \multicolumn{1}{c|}{7.97} & \multicolumn{1}{c|}{13.13} & 20.58  \\ \hline

Gray Index$^\lozenge$ \cite{GI} & \multicolumn{1}{c|}{5.18} & \multicolumn{1}{c|}{3.47} & \multicolumn{1}{c|}{4.18} & \multicolumn{1}{c|}{\secondbest 0.60} & \multicolumn{1}{c|}{12.32} & \multicolumn{1}{c|}{16.90} & 19.67  \\ \hline

APAP-LUT \cite{APAP}  & \multicolumn{1}{c|}{3.94} & \multicolumn{1}{c|}{2.82} & \multicolumn{1}{c|}{3.32} & \multicolumn{1}{c|}{0.81} & \multicolumn{1}{c|}{8.54} & \multicolumn{1}{c|}{11.90} & 16.36 \\ \hline

SIIE$^\circledcirc$ \cite{SIIE}  & \multicolumn{1}{c|}{3.51} & \multicolumn{1}{c|}{2.35} & \multicolumn{1}{c|}{2.74} & \multicolumn{1}{c|}{0.80} & \multicolumn{1}{c|}{7.56} & \multicolumn{1}{c|}{11.06} & 12.39  \\ \hline

BoCF \cite{BoCF}  & \multicolumn{1}{c|}{3.91} & \multicolumn{1}{c|}{3.31} & \multicolumn{1}{c|}{3.40} & \multicolumn{1}{c|}{1.18} & \multicolumn{1}{c|}{7.72} & \multicolumn{1}{c|}{11.24} & 12.81  \\ \hline

C4 \cite{C4}  & \multicolumn{1}{c|}{3.79} & \multicolumn{1}{c|}{2.76} & \multicolumn{1}{c|}{3.06} & \multicolumn{1}{c|}{1.05} & \multicolumn{1}{c|}{7.93} & \multicolumn{1}{c|}{11.33} & 13.39 \\ \hline

CWCC \cite{CWCC}  & \multicolumn{1}{c|}{3.71} & \multicolumn{1}{c|}{2.85} & \multicolumn{1}{c|}{3.06} & \multicolumn{1}{c|}{1.11} & \multicolumn{1}{c|}{7.79} & \multicolumn{1}{c|}{11.21} & 12.83  \\ \hline

C5$^\circledcirc$ \cite{C5}  & \multicolumn{1}{c|}{3.45} & \multicolumn{1}{c|}{2.93} & \multicolumn{1}{c|}{3.03} & \multicolumn{1}{c|}{0.95} & \multicolumn{1}{c|}{7.00} & \multicolumn{1}{c|}{10.38} & 14.01 \\ \hline

C5 (model A)  \cite{C5}  & \multicolumn{1}{c|}{3.82} & \multicolumn{1}{c|}{2.33} & \multicolumn{1}{c|}{2.81} & \multicolumn{1}{c|}{0.72} & \multicolumn{1}{c|}{9.08} & \multicolumn{1}{c|}{15.69} & 20.68  \\ \hline

C5 (model B)  \cite{C5} & \multicolumn{1}{c|}{3.30} & \multicolumn{1}{c|}{2.05} & \multicolumn{1}{c|}{2.42} & \multicolumn{1}{c|}{0.69} & \multicolumn{1}{c|}{7.58} & \multicolumn{1}{c|}{11.23} & 14.02  \\ \hline

C5 (model C)  \cite{C5} & \multicolumn{1}{c|}{3.31} & \multicolumn{1}{c|}{\secondbest 2.01} & \multicolumn{1}{c|}{2.40} & \multicolumn{1}{c|}{0.63} & \multicolumn{1}{c|}{7.67} & \multicolumn{1}{c|}{12.00} & 14.23 \\ \hline

C5 (model D)  \cite{C5}  & \multicolumn{1}{c|}{3.33} & \multicolumn{1}{c|}{2.20} & \multicolumn{1}{c|}{2.54} & \multicolumn{1}{c|}{0.73} & \multicolumn{1}{c|}{7.53} & \multicolumn{1}{c|}{10.83} & 13.33 \\ \hline

TLCC \cite{TLCC}  & \multicolumn{1}{c|}{3.73} & \multicolumn{1}{c|}{2.89} & \multicolumn{1}{c|}{3.21} & \multicolumn{1}{c|}{0.95} & \multicolumn{1}{c|}{7.54} & \multicolumn{1}{c|}{11.55} & 14.78 \\ \hline

PCC \cite{PCC}  & \multicolumn{1}{c|}{4.37} & \multicolumn{1}{c|}{3.66} & \multicolumn{1}{c|}{3.64} & \multicolumn{1}{c|}{0.89} & \multicolumn{1}{c|}{9.14} & \multicolumn{1}{c|}{15.17} & 23.34  \\ \hdashline

EMLP ($e=2$, w/ aug) & \multicolumn{1}{c|}{3.64} & \multicolumn{1}{c|}{2.66} & \multicolumn{1}{c|}{2.97} & \multicolumn{1}{c|}{0.88} & \multicolumn{1}{c|}{7.82} & \multicolumn{1}{c|}{11.5} & 13.82 \\ \hline

EMLP ($e=4$, w/ aug) & \multicolumn{1}{c|}{3.50} & \multicolumn{1}{c|}{2.63} & \multicolumn{1}{c|}{2.78} & \multicolumn{1}{c|}{0.92} & \multicolumn{1}{c|}{7.57} & \multicolumn{1}{c|}{11.62} & 13.42 \\ \hline

EMLP-$s$ ($e=8$, w/ aug) & \multicolumn{1}{c|}{3.37} & \multicolumn{1}{c|}{2.34} & \multicolumn{1}{c|}{2.74} & \multicolumn{1}{c|}{0.67} & \multicolumn{1}{c|}{7.24} & \multicolumn{1}{c|}{10.16} & 14.10  \\ \hline

EMLP ($e=8$, w/ aug) & \multicolumn{1}{c|}{3.36} & \multicolumn{1}{c|}{2.73} & \multicolumn{1}{c|}{2.84} & \multicolumn{1}{c|}{0.76} & \multicolumn{1}{c|}{7.03} & \multicolumn{1}{c|}{9.53} & 11.85  \\ \hline

ECCC-$s$ ($e=8$) & \multicolumn{1}{c|}{3.33} & \multicolumn{1}{c|}{2.92} & \multicolumn{1}{c|}{2.92} & \multicolumn{1}{c|}{0.96} & \multicolumn{1}{c|}{6.56} & \multicolumn{1}{c|}{9.38} & 10.26 \\ \hline

ECCC-$s^2$ ($e=8$) & \multicolumn{1}{c|}{3.41} & \multicolumn{1}{c|}{2.87} & \multicolumn{1}{c|}{3.03} & \multicolumn{1}{c|}{0.81} & \multicolumn{1}{c|}{6.87} & \multicolumn{1}{c|}{9.12} & 11.03  \\ \hline

ECCC ($e=2$) & \multicolumn{1}{c|}{2.94} & \multicolumn{1}{c|}{2.10} & \multicolumn{1}{c|}{2.38} & \multicolumn{1}{c|}{0.69} & \multicolumn{1}{c|}{ 6.31} & \multicolumn{1}{c|}{\best 8.13} & \secondbest 9.25  \\ \hline
						
ECCC ($e=4$) & \multicolumn{1}{c|}{3.18} & \multicolumn{1}{c|}{2.39} & \multicolumn{1}{c|}{2.60} & \multicolumn{1}{c|}{0.81} & \multicolumn{1}{c|}{6.99} & \multicolumn{1}{c|}{10.01} & 10.56  \\ \hline

ECCC ($e=8$) & \multicolumn{1}{c|}{3.00} & \multicolumn{1}{c|}{2.31} & \multicolumn{1}{c|}{2.45} & \multicolumn{1}{c|}{0.74} & \multicolumn{1}{c|}{6.66} & \multicolumn{1}{c|}{9.85} & 12.23 \\ \hline

ECCC ($e=8$, w/ aug) & \multicolumn{1}{c|}{\secondbest 2.95} & \multicolumn{1}{c|}{2.09} & \multicolumn{1}{c|}{\secondbest 2.37} & \multicolumn{1}{c|}{0.76} & \multicolumn{1}{c|}{\best 6.06} & \multicolumn{1}{c|}{\secondbest 8.14} &  9.32  \\ \hline
EMLP + ECCC & \multicolumn{1}{c|}{\best2.91 } & \multicolumn{1}{c|}{2.33 } & \multicolumn{1}{c|}{2.44 } & \multicolumn{1}{c|}{0.72 } & \multicolumn{1}{c|}{\secondbest 6.11} & \multicolumn{1}{c|}{ 8.38} &  \best 8.93 \\ \hline

\end{tabular}
}
\vspace{-2mm}
\end{table}

The results are given in Tables \ref{tab:results-val} and \ref{tab:results-test} on the validation and testing sets, respectively. We report the mean, median, tri-mean, best 25\%, worst 25\%, worst 5\%, and max angular errors in each set. We also report the total number of parameters for other methods, including ours.

Additionally, we present the results of a set of ablation studies conducted to examine the impact of the exposure factor $e$ in both EMLP and ECCC. We also studied the impact of different color spaces of input images before computing our DEF. Specifically, we used the $rg$-chromaticity and raw RGB values (similar to \cite{Chromagenic, Chromagenic1, Chromagenic2, Chromagenic3, TWOCAMS}), in addition to our main design that uses $rgb$-chromaticity. We also studied different mapping matrices, in addition to the 3$\times$3 color matrix we used in Sec. \ref{sec:def}. Specifically, we examined using the geometric 3$\times$4 affine transformation matrix and the 3$\times$3 homography matrix. 

We report results of our EMLP with and without the covariance matrix parameters in addition to the results of ECCC with different numbers of learnable biases, $n$, and with and without the proposed bias initialization. Furthermore, we show the results of ECCC without utilizing the DEF feature. Lastly, we demonstrate the influence of different sizes of input sizes on the inference accuracy of both EMLP and ECCC. 

The results show that the best results are obtained with $e=8$. This choice is intuitively sensible, as a higher exposure factor increases the differences between the dual-exposure images, resulting in greater distinction based on scene lighting conditions. The augmentation is found to be useful for EMLP; however, we did not observe consistent improvement in the case of ECCC. This is likely because EMLP has a limited number of input features, making it more susceptible to overfitting. Thus, augmentation helps in generalization. Conversely, in the case of ECCC, the inclusion of histogram features alongside the small DEF feature suggests that the model may not consistently derive benefits from augmentation. We also noted that in ECCC, using raw RGB colors of dual-exposure images (without chromaticity conversion) tends to yield better results, while in EMLP, the chromaticity conversion tends to enhance performance (see Table \ref{tab:results-val}). Results presented in Table \ref{tab:results-test} default to $rgb$-chromaticity for EMLP and raw RGB values for ECCC.

From the results presented in Tables \ref{tab:results-val} and \ref{tab:results-test}, it is evident that our proposed DEF serves as a promising feature for guiding illuminant estimators. This is demonstrated by the performance of our lightweight EMLP model (354 parameters) compared to complex models that rely solely on raw RGB image colors (e.g., TLCC \cite{TLCC} with 32 million parameters). Moreover, incorporating DEF into ECCC reduces the maximum error by over 50\% and achieves comparable results across various evaluation metrics when compared to FFCC \cite{FFCC}. 

Since both models -- namely, EMLP and ECCC -- have a reasonable number of parameters, we can combine their predictions to create an ensemble model by averaging the predicted illuminant colors from both models. The results of this ensemble model are reported in Tables \ref{tab:results-val} and \ref{tab:results-test} under `ECCC + EMLP'. 

\begin{figure}[t]
\centering
\includegraphics[width=\linewidth]{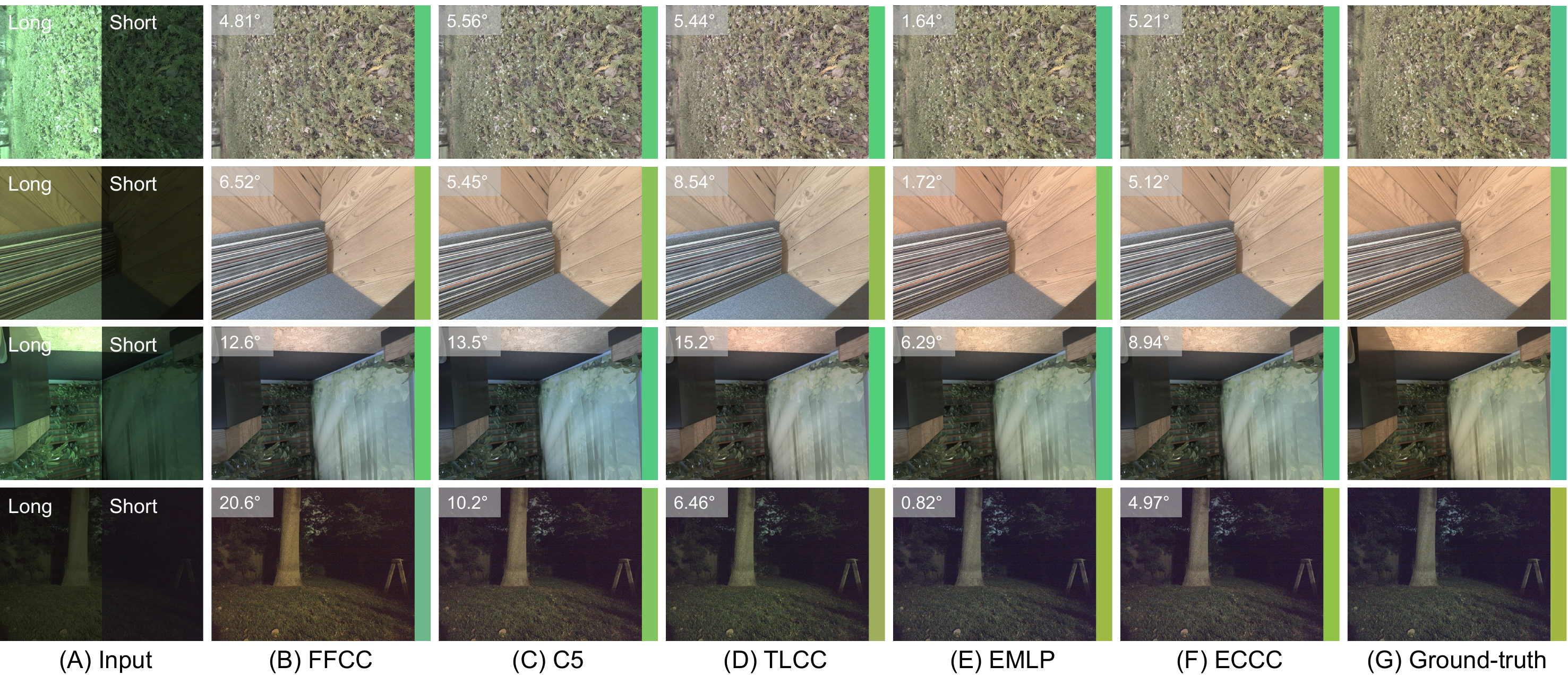}
\vspace{-3mm}
\caption{Randomly selected examples from our worst 25\% results of ECCC (top two rows are from validation set and remaining rows are from testing set). (A) Input pair of raw images captured with long and short exposures (note that other methods use a single image captured with auto exposure). (B-G) Images corrected with the estimated illuminant by: (B) FFCC \cite{FFCC}, (C) C5 \cite{C5} (chosen the best results among all variations discussed in Sec. \ref{sec:resutls}), (D) TLCC \cite{TLCC}, (E) EMLP, (F) ECCC, and (G) Ground-truth illuminant. The estimated illuminant of each method is shown on the right side of each image, along with the angular error written in the top-left corner of the image.}
\label{fig:qualitative}
\vspace{-1mm}
\end{figure}

The combined efforts of EMLP and ECCC, guided by our DEF, demonstrate promising results, outperforming several state-of-the-art approaches across various evaluation metrics while utilizing considerably fewer parameters. Qualitative examples of our results, randomly selected from the worst 25\% of ECCC results, are presented in Fig. \ref{fig:qualitative} alongside results from other methods.

\section{Conclusion and Future Work}
\label{sec:conclusion} 
In this paper, we introduced DEF, a feature derived from dual exposure images to enhance illuminant estimation. Our DEF achieves comparable results with state-of-the-art methods that employ thousands or millions of parameters (e.g., \cite{TLCC, C4}), using only 354 parameters in a straightforward MLP network. We further discuss incorporating the proposed DEF into the established CCC framework, referred to as ECCC. ECCC achieves comparable or better results on different evaluation metrics than classic CCC approaches while remaining lightweight, requiring only 6,156 parameters (50\% reduction compared to FFCC \cite{FFCC}), approximately 30 KB of memory, and running in $\sim$0.25 milliseconds per image on CPU. 

Our solution focuses on the single-camera case, intending testing on the same camera used for training. Future work includes studying the stability of this feature across different cameras. We discussed integrating the proposed DEF into the CCC framework. Further exploration may involve examining DEF's benefits for other illuminant estimation techniques that rely on convNets and raw image pixels as input. Another research direction could involve developing a spatially varying version of DEF, rather than our global feature, to utilize for spatially varying illuminant estimation and image white balancing.

\appendix
\section{Analogy to Chromagenic Color Constancy}
\label{sec:supp_discussion}

In the main paper, we draw an analogy to the chromagenic color constancy theory \cite{Chromagenic, Chromagenic1, Chromagenic2}. Our argument is grounded in the empirical findings from \cite{Chromagenic, TWOCAMS}, indicating that even when the chromagenic filter constraints are not satisfied, color mapping matrices computed to map between the colors of the main camera and a filtered/second camera still exhibit a certain degree of correlation with the scene illuminant. Practically speaking, such mapping matrices capture the color differences (or ``distortion'') between the main camera and the filtered/second camera.

Our analogy is based on the observation that, in a dual-exposure setup, the long-exposure image, $I_l$, and the short-exposure image, $I_s$, exhibit varying levels of chromatic differences and distortions based on the scene irradiance per color channel. We illustrated in the main paper the variations in the red, green, blue ratios between long and short exposure images and showed that chromatic histograms exhibit differences in similarity between the two images. We also demonstrated that these differences can vary based on the scene lighting condition.

\begin{figure}[!t]
\centering
\includegraphics[width=\linewidth]{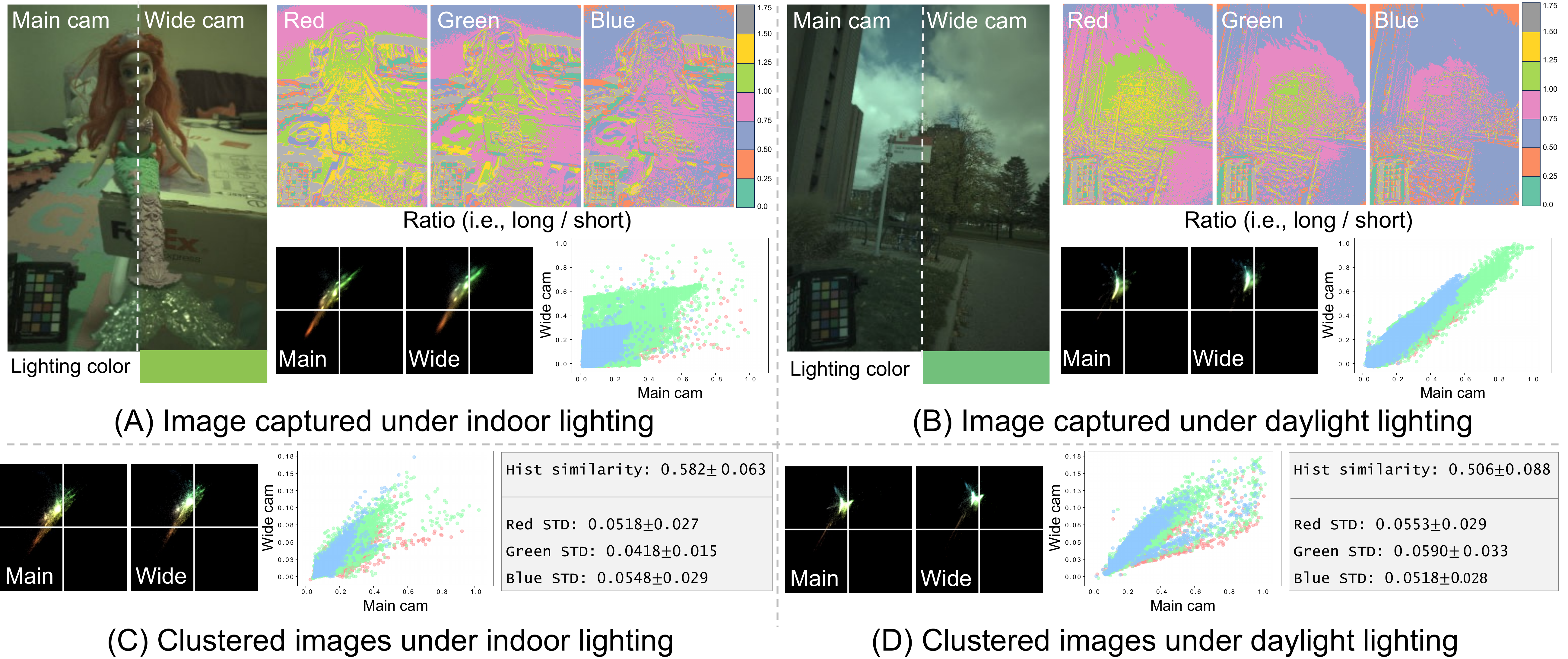}
\vspace{-3mm}
\caption{In the main paper, we drew an analogy to chromagenic color constancy, demonstrating that both cases (i.e., two cameras and dual exposure capturing) result in variations per color channel, and these differences are linked to scene irradiance and camera response function. Here, we present images captured by two cameras from \cite{TWOCAMS}. It can be observed that similar variations to the corresponding figure in the main paper in each channel occur due to the camera response function per channel. Moreover, spatial variations are noticeable based on scene irradiance and camera response function. (A) and (B) show scenes captured under indoor and outdoor lighting, respectively. In (C) and (D), we present the average $rg$-chromaticity histogram and aggregated red, green, and blue pixel values from 25 images sharing similar lighting conditions in (A) and (B), respectively.}
\label{fig:main-idea-supp}
\vspace{-1mm}
\end{figure}

In Fig. \ref{fig:main-idea-supp}, we present a similar study conducted on the two-camera dataset \cite{TWOCAMS}, which includes two cameras from a Samsung smartphone device. Comparing Fig. \ref{fig:main-idea-supp} with the corresponding figure in the main paper, we observe that both cases share a similar level of differences based on the lighting condition, albeit with less extent in the case of dual-exposure imaging. Thus, we draw our analogy by employing a 3$\times$3 color matrix that maps the $rgb$-chromaticity of $I_l$ and $I_s$ along with the covariance matrix of the ratio between each color channel in $I_s$ and $I_l$ to build our dual-exposure feature (DEF).

\section{Additional Details}
\label{sec:supp_details}

\subsection{Mapping matrices}
\label{sec:mapping_matrices}

Our DEF employs a 3$\times$3 matrix that maps between the $rgb$-chromaticity values of images $I_s$ and $I_l$. In the main paper, we presented ablation studies that utilized different mapping matrices between the chromaticity values of $I_s$ and $I_l$. Specifically, we evaluated using the geometric affine transformation instead of the linear mapping matrix. Here, we use $I^{\nu}_s$ and $I^{\nu}_l$ to refer to the $rgb$-chromaticity of the long and short-exposure images. The affine transformation matrix, denoted as $M$, between $I^{\nu}_s$ and $I^{\nu}_l$, after appending an additional constant 1 to the $rgb$-chroma triples, can be computed as follows:

\begin{equation}
\label{eq:transformation}
M = \begin{bmatrix} 
\alpha R_{\text{aff.}} & T_{\text{aff.}} \\
\mathbf{0}  & 1
\end{bmatrix}
\end{equation}

\begin{equation}
\label{eq:transformation-scale}
T_{\text{aff.}} = \text{centroid}(I^{\nu}_s) - 2 \text{ centroid}(I^{\nu}_l)
\end{equation}

\begin{equation}
\label{eq:transformation-scale}
\alpha  = \left\|I^{\nu'}_l\right\| / \left\|I^{\nu'}_s\right\|,
\end{equation}

\begin{equation}
\label{eq:transformation-rotate}
R_{\text{aff.}} = U V^T,
\end{equation}

\noindent where $I^{\nu'}_s$ and $I^{\nu'}_l$ refer to centered values of $I^{\nu}_s$ and $I^{\nu}_l$ obtained by subtracting the centroids of $I^{\nu}_s$ and $I^{\nu}_l$, respectively. $\mathbf{0}   \in \mathbb{R}^3$ is a zero vector, $U$ and $V$ are $3 \times 3$ matrices, and $S$ is a 3$\times$3 diagonal matrix. $U$, $S$, and $V$ can be obtained via singular value decomposition of the matrix multiplication of $I^{\nu'}_s$ and ${I^{\nu'}_l}^T$. Since the last row of $M$ is fixed, we excluded it from the color matrix, $C_c$ used in our DEF.

We also explored the use of a 3$\times$3 homography matrix as an alternative to the 3$\times$3 linear mapping matrix discussed in the main design of our method. Homography mapping has demonstrated its utility in various color applications \cite{finlayson2016color, finlayson2017color}. The homography matrix is computed to map between $[r, g, 1]^T$ $rg$-chromaticity values of long and short-exposure images. Based on our results, the linear transformation outperforms both geometric transformation and homography mapping.

\subsection{Exposure-Based Convolutional Color Constancy}
\label{sec:eccc-supp}

\begin{figure}[!t]
\centering
\includegraphics[width=\linewidth]{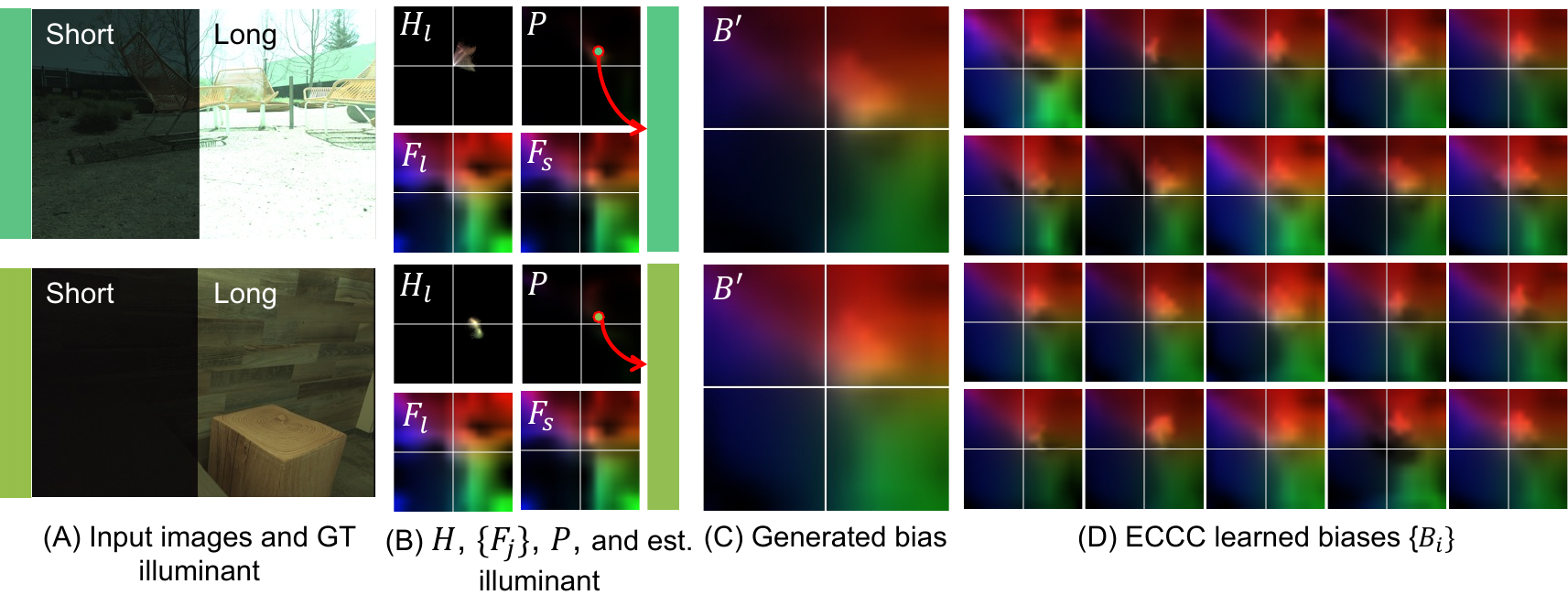}
\vspace{-3mm}
\caption{This figure shows the generated bias map alongside the learned filters of the ECCC. In (A), we show a pair of input raw images captured with long and short exposure times, along with the ground-truth illuminant color. In (B), we show the histogram of image taken with long exposure (noting that our design employs the histograms of both short and long exposure images), the learned global filters ${F_j}$ ($j \in {l, s}$), the probability map $P$, and the estimated illuminant color based on $P$. In (C), the generated bias is shown. (D) demonstrates the ECCC learned bias filters that are linearly interpolated based on the produced weights of the MLP using the input DEF associated with each pair of images.}
\label{fig:generated_bias}
\vspace{-1mm}
\end{figure}

In the main paper, we discussed a modification to the existing convolutional color constancy (CCC) framework by incorporating our DEF. The DEF is processed by a lightweight multilayer perceptron (MLP) that produces weighting factors to linearly interpolate between a set of learnable biases, generating DEF-based biases for use in the CCC. We referred to this modified version of CCC as exposured-based CCC, or ECCC for short. In our experiments we used a 64$\times$64 histogram (also we presented an ablation study on using ECCC with  32$\times$32 histograms) for ECCC and other CCC methods \cite{FFCC, C5}. The histogram, $H$, is computed as described in the following equation:

\begin{equation}
\label{eq:hist}
H(u, v) = \sum_{t=1}^{k} \left\|I^{(t)}\right\| \left[|u_t - u| \leq \varepsilon \land |v_t - v| \leq \varepsilon\right],
\end{equation}

\noindent where $k$ refers to the total number of pixels in the image, $\varepsilon = (b_{\text{max}} - b_{\text{min}})/h$, with $b_{\text{max}} = 2.85$ and $b_{\text{min}} = -2.85$ as the histogram boundary values. In ECCC, in contrast to FFCC \cite{FFCC} and C5 \cite{C5}, only colors from long-exposure and short-exposure images ($I_l$ and $I_s$) are used to create histograms, excluding edge color histograms for simplicity. Specifically, we utilized two histograms, $H_l$ and $H_s$, which represent the $uv$ chroma values of $I_l$ and $I_s$, respectively, and thus, two convolutional filters, $F_l$ and $F_s$, were learned in ECCC. Similar to FFCC and C5, FFTs are employed when convolving $F_l$ and $F_s$ over $H_l$ and $H_s$, respectively.

To train ECCC, we used additional smoothness loss terms to encourage smoothness in the learned filters and biases. These smoothness terms can be described as follows:

\begin{equation}
\label{eq:smoothness_b}
S_B\left(B\right) = \lambda_B \left(\left\|B'_{\text{up}} * \delta_u \right\|^2 + \left\|B'_{\text{up}} * \delta_v \right\|^2\right),
\end{equation}

\begin{equation}
\label{eq:smoothness_f}
S_F\left(\{F_j\}\right) = \lambda_F \sum_j{\left(\left\|\uparrow\left(F_j\right) * \delta_u \right\|^2 + \left\|\uparrow\left(F_j\right) * \delta_v \right\|^2\right)},
\end{equation}

\noindent where $\delta_u$ and $\delta_v$ are 3$\times$3 horizontal and vertical Sobel filters, respectively, and $\lambda_B=0.01$ and $\lambda_F=0.02$ are hyperparameters to control the strength of smoothness loss terms.

Figure \ref{fig:generated_bias} shows two examples of generated biases alongside the learned $n$ biases (with $n=20$). The figure also displays the learned convolutional filters for both histograms of images captured with long and short exposures ($I_l$ and $I_s$). The convolutional filters ($F_l$ and $F_s$) remain fixed in the model, while the bias dynamically changes based on the input DEF.

\subsection{Dataset}
\label{sec:data-supp}

\begin{figure}[!t]
\centering
\includegraphics[width=\linewidth]{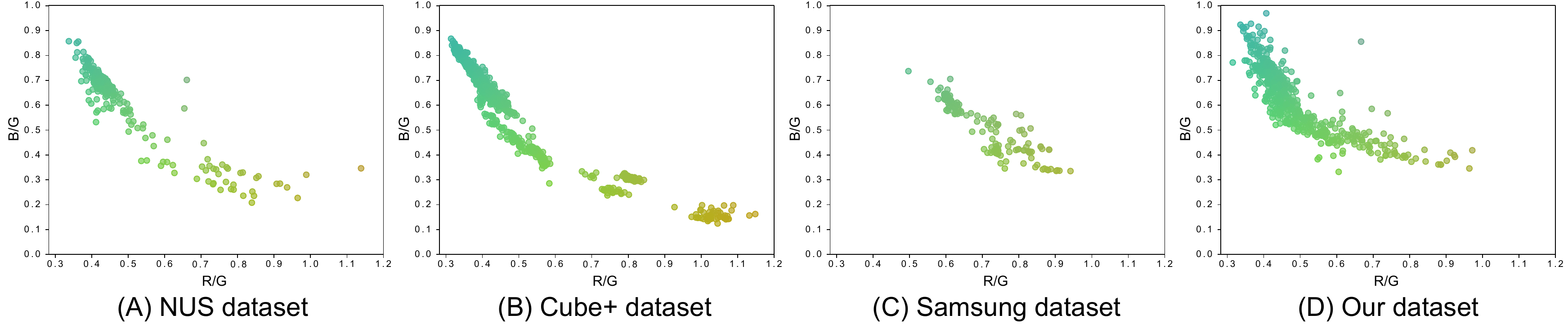}
\vspace{-3mm}
\caption{Ground truth illuminant colors of the dataset used in our paper and other datasets. (A) NUS dataset \cite{NUS}. (B) Cube+ dataset \cite{Cube+}. (C) Samsung dataset \cite{TWOCAMS}. (D) Our dataset. For the NUS and Samsung datasets, we display the ground truth from a single camera: Canon EOS-1Ds for NUS and the main camera for Samsung.}
\vspace{-1mm}
\label{fig:illum-supp}
\end{figure}

\begin{figure}[t]
\centering
\includegraphics[width=\linewidth]{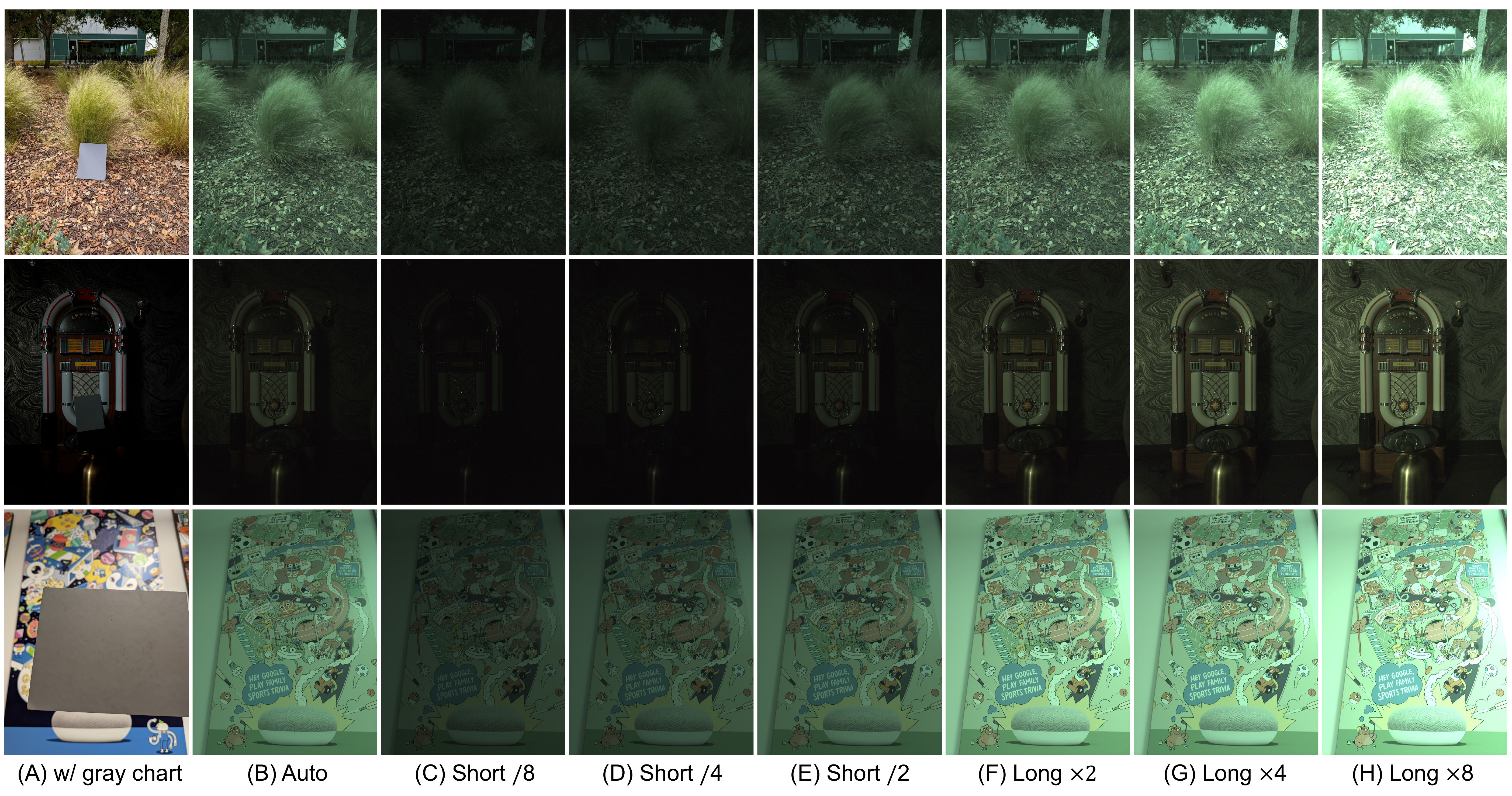}
\vspace{-3mm}
\caption{Additional examples from the dataset used in this work. For each scene, we captured the scene with a gray calibration object placed in the scene to obtain the ground-truth illuminant (A) and captured the scene using different exposure settings without the gray object (B-H). The terms `short $/ e$' (C-E) and `long $\times e$' (F-H) refer to multiplying and dividing auto exposure time by a factor $e$, respectively. The first image in (A) is displayed in sRGB, while the rest are shown in raw RGB space.}
\vspace{-1mm}
\label{fig:dataset-supp}
\end{figure}

As discussed in the main paper, we collected a dataset of multi-exposure raw images with ground-truth illuminant colors for training and evaluating our method. Figure \ref{fig:illum-supp} illustrates the distribution of R/G and B/G values for the ground-truth illuminant colors in the collected dataset. We also present illuminant distributions from other datasets (NUS \cite{NUS}, Cube+ \cite{Cube+}, and Samsung \cite{TWOCAMS}). Our dataset exhibits reasonable diversity, sometimes better, as observed when comparing with the Samsung dataset \cite{TWOCAMS}. Notably, our dataset does not lack examples for certain regions in the Planckian-like curve, unlike the NUS and Cube+ datasets \cite{NUS, Cube+}. Additional example images from our dataset are shown in Fig. \ref{fig:dataset-supp}.

\bibliographystyle{splncs04}
\bibliography{main}
\end{document}